\documentclass{article}
\usepackage{times}
\usepackage{amsmath,amssymb,amsfonts,bm}
\usepackage{graphicx,subfigure}
\usepackage{multirow,makecell,url,natbib,hyperref}
\usepackage{algorithm,algorithmic}
\usepackage[accepted]{icml2015}
\newtheorem{theorem}{Theorem}

\icmltitlerunning{Learning Transferable Features with Deep Adaptation Networks}

\begin{document}

\twocolumn[
\icmltitle{Learning Transferable Features with Deep Adaptation Networks}

\icmlauthor{Mingsheng Long$^{\dag\sharp}$}{mingsheng@tsinghua.edu.cn}
\icmlauthor{Yue Cao$^{\dag}$}{yue-cao14@mails.tsinghua.edu.cn}
\icmlauthor{Jianmin Wang$^{\dag}$}{jimwang@tsinghua.edu.cn}
\icmlauthor{Michael I. Jordan$^{\sharp}$}{jordan@berkeley.edu}
\icmladdress{$^\dag$School of Software, TNList Lab for Info. Sci. \& Tech., Institute for Data Science, Tsinghua University, China \\
$^\sharp$Department of Electrical Engineering and Computer Science, University of California, Berkeley, CA, USA}

\icmlkeywords{Deep learning, domain adaptation, two-sample test}

\vskip 0.3in
]

\begin{abstract}
Recent studies reveal that a deep neural network can learn transferable features which generalize well to novel tasks for domain adaptation. However, as deep features eventually transition from general to specific along the network, the feature transferability drops significantly in higher layers with increasing domain discrepancy. Hence, it is important to formally reduce the dataset bias and enhance the transferability in task-specific layers. In this paper, we propose a new Deep Adaptation Network (DAN) architecture, which generalizes deep convolutional neural network to the domain adaptation scenario. In DAN, hidden representations of all task-specific layers are embedded in a reproducing kernel Hilbert space where the mean embeddings of different domain distributions can be explicitly matched. The domain discrepancy is further reduced using an optimal multi-kernel selection method for mean embedding matching. DAN can learn transferable features with statistical guarantees, and can scale linearly by unbiased estimate of kernel embedding. Extensive empirical evidence shows that the proposed architecture yields state-of-the-art image classification error rates on standard domain adaptation benchmarks.
\end{abstract}

\section{Introduction}
The generalization error of supervised learning machines with limited training samples will be unsatisfactorily large, while manual labeling of sufficient training data for diverse application domains may be prohibitive. Therefore, there is incentive to establishing effective algorithms to reduce the labeling cost, typically by leveraging off-the-shelf labeled data from relevant source domains to the target domains. Domain adaptation addresses the problem that we have data from two related domains but under different distributions. The domain discrepancy poses a major obstacle in adapting predictive models across domains. For example, an object recognition model trained on manually annotated images may not generalize well on testing images under substantial variations in the pose, occlusion, or illumination. Domain adaptation establishes knowledge transfer from the labeled source domain to the unlabeled target domain by exploring domain-invariant structures that bridge different domains of substantial distribution discrepancy \cite{cite:TKDE10TLSurvey}.

One of the main approaches to establishing knowledge transfer is to learn domain-invariant models from data, which can bridge the source and target domains in an isomorphic latent feature space. In this direction, a fruitful line of prior work has focused on learning shallow features by jointly minimizing a distance metric of domain discrepancy \cite{cite:TNN11TCA,cite:ICCV13JDA,cite:ICCV13DIP,cite:ICML13Landmark,cite:ICML13TCS,cite:Arxiv14DANN,cite:NIPS14FTL}. However, recent studies have shown that deep neural networks can learn more transferable features for domain adaptation \cite{cite:ICML11DADL,cite:ICML14DeCAF,cite:NIPS14CNN}, which produce breakthrough results on some domain adaptation datasets. Deep neural networks are able to disentangle exploratory factors of variations underlying the data samples, and group features hierarchically in accordance with their relatedness to invariant factors, making representations robust to noise.

While deep neural networks are more powerful for learning general and transferable features, the latest findings also reveal that the deep features must eventually transition from general to specific along the network, and feature transferability drops significantly in higher layers with increasing domain discrepancy. In other words, the features computed in higher layers of the network must depend greatly on the specific dataset and task \cite{cite:NIPS14CNN}, which are task-specific features and are not safely transferable to novel tasks. Another curious phenomenon is that disentangling the variational factors in higher layers of the network may enlarge the domain discrepancy, as different domains with the new deep representations become more ``compact'' and are more mutually distinguishable \cite{cite:ICML11DADL}. Although deep features are salient for discrimination, enlarged dataset bias may deteriorate domain adaptation performance, resulting in statistically \emph{unbounded} risk for the target tasks \cite{cite:COLT09DAT,cite:ML10DAT}.

Inspired by the literature's latest understanding about the transferability of deep neural networks, we propose in this paper a new Deep Adaptation Network (DAN) architecture, which generalizes deep convolutional neural network to the domain adaptation scenario. The main idea of this work is to enhance the feature transferability in the task-specific layers of the deep neural network by explicitly reducing the domain discrepancy. To establish this goal, the hidden representations of all the task-specific layers are embedded to a reproducing kernel Hilbert space where the mean embeddings of different domain distributions can be explicitly matched. As mean embedding matching is sensitive to the kernel choices, an optimal multi-kernel selection procedure is devised to further reduce the domain discrepancy. In addition, we implement a linear-time unbiased estimate of the kernel mean embedding to enable scalable training, which is very desirable for deep learning. Finally, as deep models pre-trained with large-scale repositories such as ImageNet \cite{cite:Arxiv14ImageNet} are representative for general-purpose tasks \cite{cite:NIPS14CNN,cite:NIPS14LSDA}, the proposed DAN model is trained by fine-tuning from the AlexNet model \cite{cite:NIPS12CNN} pre-trained on ImageNet, which is implemented in Caffe \cite{cite:MM14Caffe}. Comprehensive empirical evidence demonstrates that the proposed architecture outperforms state-of-the-art results evaluated on the standard domain adaptation benchmarks.

The contributions of this paper are summarized as follows. (1) We propose a novel deep neural network architecture for domain adaptation, in which \emph{all} the layers corresponding to task-specific features are adapted in a layerwise manner, hence benefiting from ``deep adaptation.'' (2) We explore \emph{multiple} kernels for adapting deep representations, which substantially enhances adaptation effectiveness compared to single kernel methods. Our model can yield unbiased deep features with statistical guarantees.

\section{Related Work}\label{section:RelatedWork}
A related literature is transfer learning \cite{cite:TKDE10TLSurvey}, which builds models that bridge different domains or tasks, explicitly taking domain discrepancy into consideration. Transfer learning aims to mitigate the effort of manual labeling for machine learning~\cite{cite:TNN11TCA,cite:ICML13Landmark,cite:ICML13TCS,cite:NIPS14FTL} and computer vision \cite{cite:ECCV10SGF,cite:CVPR12GFK,cite:ICCV13DIP,cite:ICCV13JDA}, etc. It is widely recognized that the domain discrepancy in the probability distributions of different domains should be formally measured and reduced. The major bottleneck is how to match different domain distributions effectively. Most existing methods learn a new shallow representation model in which the domain discrepancy can be explicitly reduced. However, without learning deep features which can suppress domain-specific factors, the transferability of shallow features could be limited by the task-specific variability.

Deep neural networks learn nonlinear representations that disentangle and hide different explanatory factors of variation behind data samples \cite{cite:TPAMI13DLSurvey}. The learned deep representations manifest invariant factors underlying different populations and are transferable from the original tasks to similar novel tasks \cite{cite:NIPS14CNN}. Hence, deep neural networks have been explored for domain adaptation \cite{cite:ICML11DADL,cite:ICML12mSDA}, multimodal and multi-source learning problems \cite{cite:ICML11MDL,cite:KDD13MSDBN}, where significant performance gains have been obtained. However, all these methods depend on the assumption that deep neural networks can learn invariant representations that are transferable across different tasks. In reality, the domain discrepancy can be alleviated, but not removed, by deep neural networks \cite{cite:ICML11DADL}. Dataset shift has posed a bottleneck to the transferability of deep networks, resulting in statistically \emph{unbounded} risk for target tasks \cite{cite:COLT09DAT,cite:ML10DAT}.

Our work is primarily motivated by Yosinski et al.~\yrcite{cite:NIPS14CNN}, which comprehensively explores feature transferability of deep convolutional neural networks. The method focuses on a different scenario where the learning tasks are different  across domains, hence it requires sufficient target labeled examples such that the source network can be fine-tuned to the target task. In many real problems, labeled data is usually limited especially for a novel target task, hence the method cannot be directly applicable to domain adaptation. 
There are several very recent efforts in learning domain-invariant features in the context of shallow neural networks \cite{cite:NIPS14DANN,cite:Arxiv14DANN}. Due to the limited capacity of shallow architectures, the performance of these proposals does not surpass deep CNN \cite{cite:NIPS12CNN}. Tzeng et al.~\yrcite{cite:Arxiv14DDC} proposed a DDC model that adds an adaptation layer and a dataset shift loss to the deep CNN for learning a domain-invariant representation. While performance was improved, DDC only adapts a single layer of the network, which may be restrictive in that there are multiple layers where the hidden features are not transferable \cite{cite:NIPS14CNN}. DDC is also limited by suboptimal kernel matching of probability distributions \cite{cite:NIPS12MKMMD} and its quadratic computational cost that restricts transferability and scalability.

\section{Deep Adaptation Networks}\label{section:DAN}
In unsupervised domain adaptation, we are given a \emph{source} domain $\mathcal{D}_s = \{(\mathbf{x}_i^s,y^s_i)\}_{i=1}^{n_s}$ with $n_s$ labeled examples, and a \emph{target} domain $\mathcal{D}_t = \{\mathbf{x}_j^t\}_{j=1}^{n_t}$ with $n_t$ unlabeled examples. The source domain and target domain are characterized by probability distributions $p$ and $q$, respectively. We aim to construct a deep neural network which is able to learn transferable features that bridge the cross-domain discrepancy, and build a classifier $y = \theta (\mathbf{x})$ which can minimize target risk ${\epsilon_t}\left( \theta  \right) = {\Pr _{\left( {{\mathbf{x}},y} \right) \sim q}}\left[ {\theta \left( {\mathbf{x}} \right) \ne y} \right]$ using source supervision. In semi-supervised adaptation where the target has a small number of labeled examples, we denote by $\mathcal{D}_a = \{(\mathbf{x}_i^a,y_i^a)\}$ the $n_a$ annotated examples of source and target domains.

\subsection{Model}
\textbf{MK-MMD} \;
Domain adaptation is challenging in that the target domain has no (or only limited) labeled information. To approach this problem, many existing methods aim to bound the target error by the source error plus a discrepancy metric between the source and the target \cite{cite:ML10DAT}. Two classes of statistics have been explored for the \emph{two-sample} testing, where acceptance or rejection decisions are made for a null hypothesis $p=q$, given samples generated respectively from $p$ and $q$: \emph{energy distances} and \emph{maximum mean discrepancies} (MMD)~\cite{cite:AS13MMD}. In this paper, we focus on the multiple kernel variant of MMD (MK-MMD) proposed by Gretton et al.~\yrcite{cite:NIPS12MKMMD}, which is formalized to jointly maximize the two-sample test power and minimize the Type II error, i.e., the failure of rejecting a false null hypothesis.

Denote by $\mathcal{H}_k$ be the reproducing kernel Hilbert space (RKHS) endowed with a characteristic kernel $k$. The \emph{mean embedding} of distribution $p$ in $\mathcal{H}_k$ is a unique element $\mu_k(p)$ such that ${{\mathbf{E}}_{{\mathbf{x}}\sim p}}f\left( {\mathbf{x}} \right) = {\left\langle {f\left( {\mathbf{x}} \right),{\mu _k}\left( p \right)} \right\rangle _{{\mathcal{H}_k}}}$ for all $f \in \mathcal{H}_k$. The MK-MMD $d_k\left( {p,q} \right)$ between probability distributions $p$ and $q$ is defined as the RKHS distance between the mean embeddings of $p$ and $q$. The squared formulation of MK-MMD is defined as
\begin{equation}\label{eqn:MMD}
	d_k^2\left( {p,q} \right) \triangleq \left\| {{{\mathbf{E}}_p}\left[ {\phi \left( {{{\mathbf{x}}^s}} \right)} \right] - {{\mathbf{E}}_q}\left[ {\phi \left( {{{\mathbf{x}}^t}} \right)} \right]} \right\|_{{\mathcal{H}_k}}^2.
\end{equation}
The most important property is that $p=q$ iff $d_k^2\left( {p,q} \right) = 0$ \cite{cite:JMLR12MMD}. The characteristic kernel associated with the feature map $\phi$, $k\left( {{{\mathbf{x}}^s},{{\mathbf{x}}^t}} \right) = \left\langle {\phi \left( {{{\mathbf{x}}^s}} \right),\phi \left( {{{\mathbf{x}}^t}} \right)} \right\rangle $, is defined as the convex combination of $m$ PSD kernels $\{k_u\}$,
\begin{equation}\label{eqn:MK}
	\mathcal{K} \triangleq \left\{ {k = \sum\limits_{u = 1}^m {{\beta _u}{k_u}} :\sum\limits_{u = 1}^m {{\beta _u}}  = 1,{\beta _u} \geqslant 0}, {\forall u} \right\},
\end{equation}
where the constraints on coefficients $\{\beta_u\}$ are imposed to guarantee that the derived multi-kernel $k$ is characteristic. As studied theoretically in Gretton et al.~\yrcite{cite:NIPS12MKMMD}, the kernel adopted for the mean embeddings of $p$ and $q$ is critical to ensure the test power and low test error. The multi-kernel $k$ can leverage different kernels to enhance MK-MMD test, leading to a principled method for optimal kernel selection.

One of the feasible strategies for controlling the domain discrepancy is to find an abstract feature representation through which the source and target domains are similar~\cite{cite:ML10DAT}. Although this idea has been explored in several papers~\cite{cite:TNN11TCA,cite:ICML13TCS,cite:NIPS14FTL}, to date there has been no attempt to enhance the transferability of feature representation via MK-MMD in deep neural networks.

\begin{figure}[tbp]
  \centering
  \includegraphics[width=1.0\columnwidth]{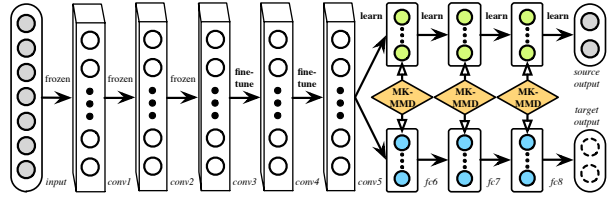}
  \caption{The DAN architecture for learning transferable features. Since deep features eventually transition from general to specific along the network, (1) the features extracted by convolutional layers $conv1$--$conv3$ are general, hence these layers are frozen, (2) the features extracted by layers $conv4$--$conv5$ are slightly less transferable, hence these layers are learned via fine-tuning, and (3) fully connected layers $fc6$--$fc8$ are tailored to fit specific tasks, hence they are not transferable and should be adapted with MK-MMD.}
  \label{fig:DAN}
\end{figure}

\textbf{Deep Adaptation Networks (DAN)} \;
In this paper, we explore the idea of MK-MMD-based adaptation for learning transferable features in deep networks. We start with deep convolutional neural networks (CNN) \cite{cite:NIPS12CNN}, a strong model when it is adapted to novel tasks \cite{cite:ICML14DeCAF,cite:NIPS14LSDA}. The main challenge is that the target domain has no or just limited labeled information, hence directly adapting CNN to the target domain via fine-tuning is impossible or is prone to over-fitting. With the idea of domain adaptation, we are targeting a deep adaptation network (DAN) that can exploit both source-labeled data and target-unlabeled data. Figure~\ref{fig:DAN} gives an illustration of the proposed DAN model.

We extend the AlexNet architecture \cite{cite:NIPS12CNN}, which is comprised of five convolutional layers ($conv1$--$conv5$) and three fully connected layers ($fc6$--$fc8$). Each $fc$ layer $\ell$ learns a nonlinear mapping ${\mathbf{h}}_i^\ell  = {f^\ell }\left( {{{\mathbf{W}}^{\ell}}{\mathbf{h}}_i^{\ell  - 1} + {{\mathbf{b}}^{\ell}}} \right)$, where $\mathbf{h}_i^\ell$ is the $\ell$th layer hidden representation of point $\mathbf{x}_i$, $\mathbf{W}^\ell$ and $\mathbf{b}^\ell$ are the weights and bias of the $\ell$th layer, and $f^\ell$ is the activation, taking as rectifier units $f^\ell(\mathbf{x}) = \max(\mathbf{0}, \mathbf{x})$ for hidden layers or softmax units ${f^\ell }\left( {\mathbf{x}} \right) = {e^{\mathbf{x}}}/\sum\nolimits_{j = 1}^{\left| {\mathbf{x}} \right|} {{e^{{x_j}}}} $ for the output layer. Letting $\Theta  = \left\{ {{{\mathbf{W}}^\ell },{{\mathbf{b}}^\ell }} \right\}_{\ell  = 1}^l$ denote the set of all CNN parameters, the empirical risk of CNN is
\begin{equation}\label{eqn:CNN}
	\mathop {\min }\limits_\Theta  \frac{1}{{{n_a}}}\sum\limits_{i = 1}^{{n_a}} {J\left( {\theta \left( {{\mathbf{x}}_i^a} \right),y_i^a} \right)},
\end{equation}
where $J$ is the cross-entropy loss function, and ${\theta \left( {{\mathbf{x}}_i^a} \right)}$ is the conditional probability that the CNN assigns $\mathbf{x}_i^a$ to label $y_i^a$. We will not discuss how to compute the convolutional layers as we will not impose distribution-adaptation regularization in those layers, given that the convolutional layers can learn generic features that tend to be transferable in layers $conv1$--$conv3$ and are slightly domain-biased in $conv4$--$conv5$~\cite{cite:NIPS14CNN}.  Hence, when adapting the pre-trained AlexNet to the target, we opt to freeze $conv1$--$conv3$ and fine-tune $conv4$--$conv5$ to preserve the efficacy of fragile co-adaptation \cite{cite:Arxiv12CoA}.

In standard CNNs, deep features must eventually transition from general to specific by the last layer of the network, and the transferability gap grows with the domain discrepancy and becomes particularly large when transferring the higher layers $fc6$--$fc8$ \cite{cite:NIPS14CNN}. In other words, the $fc$ layers are tailored to their original task at the expense of degraded performance on the target task, hence they cannot be directly transferred to the target domain via fine-tuning with limited target supervision. In this paper, we fine-tune CNN on the source labeled examples and require the distributions of the source and target to become similar under the hidden representations of fully connected layers $fc6$--$fc8$. This can be realized by adding an MK-MMD-based multi-layer adaptation regularizer \eqref{eqn:MMD} to the CNN risk \eqref{eqn:CNN}:
\begin{equation}\label{eqn:DAN}
	\mathop {\min }\limits_\Theta  \frac{1}{{{n_a}}}\sum\limits_{i = 1}^{{n_a}} {J\left( {\theta \left( {{\mathbf{x}}_i^a} \right),y_i^a} \right)}  + \lambda \sum\limits_{\ell  = {l _1}}^{{l _2}} {d_k^2\left( {\mathcal{D}_s^\ell ,\mathcal{D}_t^\ell } \right)},
\end{equation}
where $\lambda > 0$ is a penalty parameter, $l_1$ and $l_2$ are layer indices between which the regularizer is effective. In our implementation of DAN, we set $l_1 = 6$ and $l_2 = 8$, although different configurations are also possible, depending on the size of the labeled source dataset and the number of parameters in the layers that are to be fine-tuned. $\mathcal{D}_ * ^\ell  = \left\{ {{\mathbf{h}}_i^{ * \ell }} \right\}$ is the $\ell$th layer hidden representation for the source and target examples, and $d_k^2\left( {\mathcal{D}_s^\ell ,\mathcal{D}_t^\ell } \right)$ is the MK-MMD between the source and target evaluated on the $\ell$th layer representation.

Training a deep CNN requires a large amount of labeled data, which is prohibitive for many domain adaptation problems, hence we start with an AlexNet model pre-trained on ImageNet 2012 and fine-tune it as in Yosinski et al.~\yrcite{cite:NIPS14CNN}. With the proposed DAN optimization framework \eqref{eqn:DAN}, we are able to learn transferable features from a source domain to a related target domain. The learned representation can both be salient benefiting from CNN, and unbiased thanks to MK-MMD. Two important advantages that distinguish DAN from relevant literature are: (1) \emph{multi-layer} adaptation. As revealed by \cite{cite:NIPS14CNN}, feature transferability gets worse on $conv4$--$conv5$ and significantly drops on $fc6$--$fc8$, hence it is critical to adapt multiple layers instead of only one layer. In other words, adapting a single layer cannot undo the dataset bias between the source and the target, since there are other layers that are not transferable. Another benefit of multi-layer adaptation is that by jointly adapting the representation layers and the classifier layer, we could essentially bridge the domain discrepancy underlying \emph{both} the marginal distribution and the conditional distribution, which is crucial for domain adaptation \cite{cite:ICML13TCS}. (2) \emph{multi-kernel} adaptation. As pointed out by Gretton et al.~\yrcite{cite:NIPS12MKMMD}, kernel choice is critical to the testing power of MMD since different kernels may embed probability distributions in different RKHSs where different orders of sufficient statistics can be emphasized. This is crucial for moment matching, which is not well explored by previous domain adaptation methods.

\subsection{Algorithm}
\textbf{Learning $\Theta$} \;
Using the kernel trick, MK-MMD \eqref{eqn:MMD} can be computed as the expectation of kernel functions $d_k^2\left( {p,q} \right) = {{\mathbf{E}}_{{{\mathbf{x}}^s}{{{\mathbf{x'}}}^s}}}k({{\mathbf{x}}^s},{{\mathbf{x'}}^s}) + {{\mathbf{E}}_{{{\mathbf{x}}^t}{{{\mathbf{x'}}}^t}}}k({{\mathbf{x}}^t},{{\mathbf{x'}}^t}) - 2{{\mathbf{E}}_{{{\mathbf{x}}^s}{{\mathbf{x}}^t}}}k({{\mathbf{x}}^s},{{\mathbf{x}}^t})$, where ${{\mathbf{x}}^s},{{{\mathbf{x'}}}^s}\mathop \sim \limits^{iid} p$, ${{\mathbf{x}}^t},{{{\mathbf{x'}}}^t}\mathop \sim \limits^{iid} q$, and $k \in \mathcal{K}$. However, this computation incurs a complexity of $O(n^2)$, which is rather undesirable for deep CNNs, as the power of deep neural networks largely derives from learning with large-scale datasets. Moreover, the summation over pairwise similarities between data points makes mini-batch stochastic gradient descent (SGD) more difficult, whereas mini-batch SGD is crucial to the training effectiveness of deep networks. While prior work based on MMD \cite{cite:TNN11TCA,cite:Arxiv14DDC} rarely addresses this issue, we believe it is critical in the context of deep learning. In this paper, we adopt the unbiased estimate of MK-MMD \cite{cite:NIPS12MKMMD} which can be computed with linear complexity. More specifically, $d_k^2\left( {p,q} \right) = \frac{2}{{{n_s}}}\sum\nolimits_{i = 1}^{{n_s}/2} {g_k\left( {{{\mathbf{z}}_i}} \right)} $, where we denote quad-tuple ${{\mathbf{z}}_i} \triangleq ( {{\mathbf{x}}_{2i - 1}^s,{\mathbf{x}}_{2i}^s,{\mathbf{x}}_{2i - 1}^t,{\mathbf{x}}_{2i}^t} )$, and evaluate multi-kernel function $k$ on each quad-tuple ${\bf z}_i$ by ${g_k}\left( {{{\mathbf{z}}_i}} \right) \triangleq k({\mathbf{x}}_{2i - 1}^s,{\mathbf{x}}_{2i}^s) + k({\mathbf{x}}_{2i - 1}^t,{\mathbf{x}}_{2i}^t) - k({\mathbf{x}}_{2i - 1}^s,{\mathbf{x}}_{2i}^t) - k({\mathbf{x}}_{2i}^s,{\mathbf{x}}_{2i - 1}^t)$. This approach computes an expectation of independent variables as in \eqref{eqn:MMD} with cost $O(n)$.

When we train deep CNN by mini-batch SGD, we only need to consider the gradient of objective \eqref{eqn:DAN} with respect to each data point ${\bf x}_i$. Since the linear-time MK-MMD takes a nice summation form that can be readily decoupled into the sum of $g_k({\bf z}_i)$'s, we only need to compute the gradients $\frac{{\partial {g_k}( {{\mathbf{z}}_i^\ell } )}}{{\partial {\Theta ^\ell }}}$ for the quad-tuple ${\mathbf{z}}_i^\ell  = \left( {{\mathbf{h}}_{2i - 1}^{s\ell },{\mathbf{h}}_{2i}^{s\ell },{\mathbf{h}}_{2i - 1}^{t\ell },{\mathbf{h}}_{2i}^{t\ell }} \right)$ of the $\ell$th layer hidden representation. To be consistent with the gradient of MK-MMD, we need to compute the corresponding gradient of CNN risk $\frac{{\partial J\left( {{{\mathbf{z}}_i}} \right)}}{{\partial {{{\Theta}}^\ell }}}$, where $J\left( {{{\mathbf{z}}_i}} \right) = \sum\nolimits_{i'} {J\left( {\theta \left( {{\mathbf{x}}_{i'}^a} \right),y_{i'}^a} \right)} $, and $\{({\bf x}_{i'}^a, y_{i'}^a)\}$ indicates the labeled examples in quad-tuple ${\bf z}_i$---for instance, in unsupervised adaptation where the target domain has no labeled data, we have $\{ ({\mathbf{x}}_{i'}^a,y_{i'}^a)\}  = \{ ({\mathbf{x}}_{2i - 1}^s,y_{2i - 1}^s),({\mathbf{x}}_{2i}^s,y_{2i}^s)\}$. To perform a mini-batch update, we compute the gradient of objective \eqref{eqn:DAN} with respect to the $\ell$th layer parameter ${\Theta}^\ell$ as
\begin{equation}\label{eqn:UpdateW}
	{\nabla _{{\Theta ^\ell }}} = \frac{{\partial J\left( {{{\mathbf{z}}_i}} \right)}}{{\partial {\Theta ^\ell }}} + \lambda \frac{{\partial {g_k}\left( {{\mathbf{z}}_i^\ell } \right)}}{{\partial {\Theta ^\ell }}}.
\end{equation}
Such a mini-batch SGD can be easily implemented within the Caffe framework for CNNs~\cite{cite:MM14Caffe}. Given kernel $k$ as the linear combination of $m$ Gaussian kernels $\{ {{k_u}\left( {{{\mathbf{x}}_i},{{\mathbf{x}}_j}} \right) = {e^{ - {{\left\| {{{\mathbf{x}}_i} - {{\mathbf{x}}_j}} \right\|}^2}/\gamma _u}}} \}$, the gradient $\frac{{\partial {g_k}\left( {{\mathbf{z}}_i^\ell } \right)}}{{\partial {\Theta ^\ell }}}$ can be readily computed using the chain rule. For instance, 
\begin{equation}
	\begin{aligned}
		\frac{{\partial k({\mathbf{h}}_{2i - 1}^{s\ell },{\mathbf{h}}_{2i}^{t\ell })}}{{\partial {{\mathbf{W}}^\ell }}} &=  - \sum\limits_{u = 1}^m {\frac{{2{\beta _u}}}{{\gamma _u}}{k_u}\left( {{\mathbf{h}}_{2i - 1}^{s\ell },{\mathbf{h}}_{2i}^{t\ell }} \right)} \\
	  & \times \left( {{\mathbf{h}}_{2i - 1}^{s\ell } - {\mathbf{h}}_{2i}^{t\ell }} \right) \\
	  & \times {\left( {\mathbb{I}\left[ {{\mathbf{h}}_{2i - 1}^{s(\ell  - 1)}} \right] - \mathbb{I}\left[ {{\mathbf{h}}_{2i}^{t(\ell  - 1)}} \right]} \right)^\mathsf{T}}, \\
	\end{aligned}
\end{equation}
where the last row computes the gradient of the $\ell$th layer rectifier units, with $\mathbb{I}$ being defined as an indicator such that $\mathbb{I}\left[ {{\mathbf{h}}_{ji}^{\ell  - 1}} \right] = {\mathbf{h}}_{ji}^{\ell  - 1}$ if ${\mathbf{W}}_{ j \cdot }^\ell {\mathbf{h}}_{i}^{\ell  - 1} + {\mathbf{b}_j^{\ell}} \geqslant 0$, else $\mathbb{I}\left[ {{\mathbf{h}}_{ji}^{\ell  - 1}} \right] = 0$.

\textbf{Learning ${\bm \beta}$} \;
The proposed multi-layer adaptation regularizer performs layerwise matching by MK-MMD, hence we seek to learn optimal kernel parameter ${\bm \beta}$ for MK-MMD by jointly maximizing the test power and minimizing the Type II error \cite{cite:NIPS12MKMMD}, leading to the optimization
\begin{equation}\label{eqn:MKO}
	\mathop {\max }\limits_{k \in \mathcal{K}} d_k^2\left( {\mathcal{D}_s^\ell ,\mathcal{D}_t^\ell } \right) \sigma _{k}^{ - 2},
\end{equation}
where ${{ \sigma }^2_k} = {{\mathbf{E}}_{\mathbf{z}}}g_k^2\left( {\mathbf{z}} \right) - {\left[ {{{\mathbf{E}}_{\mathbf{z}}}{g_k}\left( {\mathbf{z}} \right)} \right]^2}$ is estimation variance. Letting ${\mathbf{d}} = {( {{d_1},{d_2}, \ldots ,{d_m}} )^{\sf T}}$, each $d_u$ is MMD via kernel $k_u$. Covariance ${\mathbf{Q}} = \operatorname{cov} \left( g_k \right) \in \mathbb{R}^{m \times m}$ can be computed in $O(m^2 n)$ cost, i.e. ${{\bf Q}_{uu'}} = \frac{4}{{{n_s}}}\sum\nolimits_{i = 1}^{{n_s}/4} {{g^\Delta_{k_u}}\left( {{{{\mathbf{\bar z}}}_i}} \right){g^\Delta_{k_{u'}}}\left( {{{{\mathbf{\bar z}}}_i}} \right)} $, where ${{{\mathbf{\bar z}}}_i} \triangleq \left( {{{\mathbf{z}}_{2i - 1}},{{\mathbf{z}}_{2i}}} \right)$ and ${g^\Delta_{k_u}}\left( {{{{\mathbf{\bar z}}}_i}} \right) \triangleq {g_{{k_u}}}\left( {{{\mathbf{z}}_{2i - 1}}} \right) - {g_{{k_u}}}\left( {{{\mathbf{z}}_{2i}}} \right)$. Hence \eqref{eqn:MKO} reduces to a quadratic program (QP),
\begin{equation}\label{eqn:UpdateBeta}
	\mathop {\min }\limits_{{{\mathbf{d}}^{\sf T}}{\bm\beta}  = 1, {\bm\beta}  \geqslant {\mathbf{0}}} {{\bm\beta} ^{\sf T}}\left( {{\mathbf{Q}} + \varepsilon {\mathbf{I}}} \right){\bm\beta},
\end{equation}
where $\varepsilon = 10^{-3}$ is a small regularizer to make the problem well-defined. By solving \eqref{eqn:UpdateBeta}, we obtain a multi-kernel $k = \sum\nolimits_{u = 1}^m {{\beta _u}{k_u}} $ that jointly maximizes the test power and minimizes the Type II error. 

We note that the DAN objective \eqref{eqn:DAN} is essentially a minimax problem; i.e., we compute $\mathop {\min }\limits_\Theta  \mathop {\max }\limits_{{\cal K}} d_k^2\left( {\mathcal{D}_s^\ell ,\mathcal{D}_t^\ell } \right)\sigma _k^{ - 2}$. The CNN parameter $\Theta$ is learned by minimizing MK-MMD as a domain discrepancy, while the MK-MMD parameter ${\bm\beta}$ is learned by minimizing the Type II error. Both criteria are dedicated to an effective adaptation of domain discrepancy, aiming to consolidate the transferability of DAN features. We accordingly adopt an alternating optimization that updates $\Theta$ by mini-batch SGD \eqref{eqn:UpdateW} and ${\bm\beta}$ by QP \eqref{eqn:UpdateBeta} iteratively. Both updates cost $O(n)$ and are scalable to large datasets.

\subsection{Analysis}
We provide an analysis of the expected target-domain risk of our approach, making use of the theory of domain adaptation~\cite{cite:NIPS07DAT,cite:ML10DAT,cite:COLT09DAT} and the theory of kernel embedding of probability distributions \cite{cite:NIPS09KCC,cite:JMLR12MMD,cite:NIPS12MKMMD}.

\begin{theorem}
	Let $\theta \in \mathcal{H}$ be a hypothesis, ${\epsilon _s}(\theta )$ and ${\epsilon _t}(\theta )$ be the expected risks of source and target respectively, then
  \begin{equation}
  	{\epsilon _t}(\theta ) \leqslant {\epsilon _s}(\theta ) + 2 {d_k}(p,q) + C,
  \end{equation}
  where $C$ is a constant for the complexity of hypothesis space and the risk of an ideal hypothesis for both domains.
\end{theorem}

\emph{Proof sketch:} A result from Ben-David et al.~\yrcite{cite:NIPS07DAT} shows that ${\epsilon _t}(\theta ) \leqslant {\epsilon _s}(\theta ) + {d_\mathcal{H}}(p,q) + {C_0}$, where ${d_\mathcal{H}}(p,q)$ is the ${\cal H}$-divergence between $p$ and $q$, which is defined as
\begin{equation}
	{d_\mathcal{H}}(p,q) \triangleq 2\mathop {\sup }\limits_{\eta  \in \mathcal{H}} \left| {\mathop {\Pr }\limits_{{{\mathbf{x}}^s} \sim p} \left[ {\eta ({{\mathbf{x}}^s}) = 1} \right] - \mathop {\Pr }\limits_{{{\mathbf{x}}^t} \sim q} \left[ {\eta ({{\mathbf{x}}^t}) = 1} \right]} \right|.
\end{equation}
The ${\cal H}$-divergence relies on the capacity of the hypothesis space ${\cal H}$ to distinguish distributions $p$ from $q$, and $\eta \in \mathcal{H}$ can be viewed as a \emph{two-sample} classifier. By choosing $\eta$ as a (kernel) Parzen window classifier \cite{cite:NIPS09KCC}, ${d_\mathcal{H}}(p,q)$ can be bounded by the empirical estimate
\begin{equation}\label{eqn:Hdiverence}
	\begin{aligned}
  	{d_\mathcal{H}}(p,q) & \leqslant {{\hat d}_\mathcal{H}}({\mathcal{D}_s},{\mathcal{D}_t}) + {{C_1}} \\
		& \leqslant \; \scriptstyle{ 2\left( {1 - \mathop {\inf }\limits_{\eta  \in \mathcal{H}} \left[ {\sum\limits_{i = 1}^{{n_s}} {\frac{{L\left[ {\eta ({\mathbf{x}}_i^s) = 1} \right]}}{{{n_s}}}}  + \sum\limits_{j = 1}^{{n_t}} {\frac{{L\left[ {\eta ({\mathbf{x}}_j^t) =  - 1} \right]}}{{{n_t}}}} } \right]} \right) + {C_1} } \\
   	& = 2\left( {1 + {d_k}(p,q)} \right) + {C_1}, \\ 
	\end{aligned}
\end{equation}
where $L(\cdot)$ is the linear loss function of the Parzen window classifier $\eta$, $L[\eta  = 1] \triangleq  - \eta ,L[\eta  =  - 1] \triangleq \eta $. By explicitly minimizing MK-MMD in multiple layers, the features and classifier learned by the proposed DAN model can decrease the upper bound on target risk. The source classifier and the two-sample classifier together provide a way to assess the adaptation performance, and can facilitate model selection. Note that we maximize MK-MMD w.r.t. ${\bm\beta}$ \eqref{eqn:MKO} to minimize Type II test error, and to help the Parzen window classifier achieve minimal risk of two-sample discrimination in \eqref{eqn:Hdiverence}.

\section{Experiments}\label{section:Experiments}
We compare the DAN model to state-of-the-art transfer learning and deep learning methods on both unsupervised and semi-supervised adaptation problems, focusing on the efficacy of multi-layer adaptation with multi-kernel MMD.

\subsection{Setup}
\textbf{Office-31} \cite{cite:ECCV10SGF} \;
This dataset is a standard benchmark for domain adaptation. It consists of 4,652 images within 31 categories collected from three distinct domains: \textit{Amazon} (\textbf{A}), which contains images downloaded from \url{amazon.com}, \textit{Webcam} (\textbf{W}) and \textit{DSLR} (\textbf{D}), which are images taken by web camera and digital SLR camera in an office with different environment variation, respectively. We evaluate our method across the 3 transfer tasks, \textbf{A} $\rightarrow$ \textbf{W}, \textbf{D} $\rightarrow$ \textbf{W} and \textbf{W} $\rightarrow$ \textbf{D}, which are commonly adopted in deep learning methods \cite{cite:ICML14DeCAF,cite:Arxiv14DDC}. For completeness, we further include the evaluation on the other 3 transfer tasks, \textbf{A} $\rightarrow$ \textbf{D}, \textbf{D} $\rightarrow$ \textbf{A} and \textbf{W} $\rightarrow$ \textbf{A}.
\textbf{Office-10 + Caltech-10} \cite{cite:CVPR12GFK}.
This dataset consists of the 10 common categories shared by the Office-31 and Caltech-256 (\textbf{C}) \cite{cite:TR07Caltech} datasets and is widely adopted in transfer learning methods \cite{cite:ICCV13JDA,cite:ICCV13DIP}. We can build another 6 transfer tasks: \textbf{A} $\rightarrow$ \textbf{C}, \textbf{W} $\rightarrow$ \textbf{C}, \textbf{D} $\rightarrow$ \textbf{C}, \textbf{C} $\rightarrow$ \textbf{A}, \textbf{C} $\rightarrow$ \textbf{W}, and \textbf{C} $\rightarrow$ \textbf{D}. With more transfer tasks, we are targeting an \emph{unbiased} look at the dataset bias \cite{cite:CVPR11DB}.

We compare to a variety of methods: TCA \cite{cite:TNN11TCA}, GFK \cite{cite:CVPR12GFK}, CNN \cite{cite:NIPS12CNN}, LapCNN \cite{cite:ICML08LapCNN}, and DDC \cite{cite:Arxiv14DDC}. Specifically, TCA is a conventional transfer learning method based on MMD-regularized PCA. GFK is a widely-adopted method for our datasets which interpolates across intermediate subspaces to bridge the source and target. CNN was the leading method in the ImageNet 2012 competition, and it turns out to be a strong model for learning transferable features \cite{cite:NIPS14CNN}. LapCNN is a semi-supervised variant of CNN based on Laplacian graph regularization. Finally, DDC is a domain adaptation variant of CNN that adds an adaptation layer between the $fc7$ and $fc8$ layers that is regularized by single-kernel MMD. We implement the CNN-based methods, i.e., CNN, LapCNN, DDC, and DAN based on the Caffe \cite{cite:MM14Caffe} implementation of AlexNet \cite{cite:NIPS12CNN} trained on the ImageNet dataset. In order to study the efficacy of \emph{multi-layer} adaptation and \emph{multi-kernel} MMD, we evaluate several variants of DAN: (1) DAN using only one hidden layer, either $fc7$ or $fc8$ for adaptation, termed DAN$_7$ and DAN$_8$ respectively; (2) DAN using single-kernel MMD for adaptation, termed DAN$_{\textrm{SK}}$.

We mainly follow standard evaluation protocol for unsupervised adaptation and use all source examples with labels and all target examples without labels \cite{cite:ICML13Landmark}. To make our results directly comparable to most published results, we report a classical protocol \cite{cite:ECCV10SGF} in that we randomly down-sample the source examples, and further require 3 labeled target examples per category for semi-supervised adaptation. We compare the averages and standard errors of classification accuracy for each task. For baseline methods, we follow the standard procedures for model selection as explained in their respective papers. For MMD-based methods (i.e., TCA, DDC, and DAN), we use a Gaussian kernel $k\left( {{{\mathbf{x}}_i},{{\mathbf{x}}_j}} \right) = {e^{ - {{\left\| {{{\mathbf{x}}_i} - {{\mathbf{x}}_j}} \right\|}^2} /\gamma }}$ with the bandwidth $\gamma$ set to the median pairwise distances on the training data---the \emph{median heuristic} \cite{cite:NIPS12MKMMD}. We use multi-kernel MMD for DAN, and consider a family of $m$ Gaussian kernels $\{ k_u \}_{u=1}^m$ by varying bandwidth $\gamma_u$ between $2^{-8}\gamma$ and $2^{8}\gamma$ with a multiplicative step-size of $2^{1/2}$ \cite{cite:NIPS12MKMMD}. As minimizing MMD is equivalent to maximizing the error of classifying the source from the target (two-sample classifier) \cite{cite:NIPS09KCC}, we can automatically select the MMD penalty parameter $\lambda$ on a validation set (comprised of source-labeled instances and target-unlabeled instances) by jointly assessing the test errors of the source classifier and the two-sample classifier.
We use the fine-tuning architecture \cite{cite:NIPS14CNN}, however, due to limited training examples in our datasets, we fix convolutional layers $conv1$--$conv3$ that were copied from pre-trained model, fine-tune $conv4$--$conv5$ and fully connected layers $fc6$--$fc7$, and train classifier layer $fc8$, both via back propagation. As the classifier is trained from scratch, we set its learning rate to be 10 times that of the lower layers. We use stochastic gradient descent (SGD) with 0.9 momentum and the learning rate annealing strategy implemented in Caffe, and cross-validate base learning rate between $10^{-5}$ and $10^{-2}$ with a multiplicative step-size $10^{1/2}$.

\begin{table*}[tbp]
    \addtolength{\tabcolsep}{1pt}
    \centering
    \caption{Accuracy on \emph{Office-31} dataset with standard unsupervised adaptation protocol \cite{cite:ICML13Landmark}.}
    \label{table:office31}
    \begin{tabular}{cccccccc}
        \Xhline{1pt}
        Method & A $\rightarrow$ W & D $\rightarrow$ W & W $\rightarrow$ D & A $\rightarrow$ D & D $\rightarrow$ A & W $\rightarrow$ A & Average \\
        \hline
				TCA & 21.5 $\pm$ 0.0 & 50.1 $\pm$ 0.0 & 58.4 $\pm$ 0.0 & 11.4 $\pm$ 0.0 & 8.0 $\pm$ 0.0 & 14.6 $\pm$ 0.0 & 27.3 \\
				GFK & 19.7 $\pm$ 0.0 & 49.7 $\pm$ 0.0 & 63.1 $\pm$ 0.0 & 10.6 $\pm$ 0.0 & 7.9 $\pm$ 0.0 & 15.8 $\pm$ 0.0 & 27.8 \\
				CNN & 61.6 $\pm$ 0.5 & 95.4 $\pm$ 0.3 & \underline{99.0} $\pm$ 0.2 & 63.8 $\pm$ 0.5 & 51.1 $\pm$ 0.6 & 49.8 $\pm$ 0.4 & 70.1 \\
				LapCNN & 60.4 $\pm$ 0.3 & 94.7 $\pm$ 0.5 & \textbf{99.1} $\pm$ 0.2 & 63.1 $\pm$ 0.6 & 51.6 $\pm$ 0.4 & 48.2 $\pm$ 0.5 & 69.5 \\
				DDC & 61.8 $\pm$ 0.4 & 95.0 $\pm$ 0.5 & 98.5 $\pm$ 0.4 & 64.4 $\pm$ 0.3 & 52.1 $\pm$ 0.8 & \underline{52.2} $\pm$ 0.4 & 70.6 \\
        \hline
				DAN$_7$ & 63.2 $\pm$ 0.2 & 94.8 $\pm$ 0.4 & 98.9 $\pm$ 0.3 & 65.2 $\pm$ 0.4 & 52.3 $\pm$ 0.4 & 52.1 $\pm$ 0.4 & 71.1 \\
				DAN$_8$ & \underline{63.8} $\pm$ 0.4 & 94.6 $\pm$ 0.5 & 98.8 $\pm$ 0.6 & 65.8 $\pm$ 0.4 & 52.8 $\pm$ 0.4 & 51.9 $\pm$ 0.5 & 71.3 \\
				DAN$_{\textrm{SK}}$ & 63.3 $\pm$ 0.3 & \underline{95.6} $\pm$ 0.2 & \underline{99.0} $\pm$ 0.4 & \underline{65.9} $\pm$ 0.7 & \underline{53.2} $\pm$ 0.5 & 52.1 $\pm$ 0.4 & \underline{71.5} \\
				DAN & \textbf{68.5} $\pm$ 0.4 & \textbf{96.0} $\pm$ 0.3 & \underline{99.0} $\pm$ 0.2 & \textbf{67.0} $\pm$ 0.4 & \textbf{54.0} $\pm$ 0.4 & \textbf{53.1} $\pm$ 0.3 & \textbf{72.9} \\
        \Xhline{1pt}
    \end{tabular}
\end{table*}

\begin{table*}[tbp]
    \addtolength{\tabcolsep}{1pt}
    \centering
    \caption{Accuracy on \emph{Office-10 + Caltech-10} dataset with standard unsupervised adaptation protocol \cite{cite:ICML13Landmark}.}
    \label{table:office10}
    \begin{tabular}{cccccccc}
        \Xhline{1pt}
        Method & A $\rightarrow$ C & W $\rightarrow$ C & D $\rightarrow$ C & C $\rightarrow$ A & C $\rightarrow$ W & C $\rightarrow$ D & Average \\
        \hline
				TCA & 42.7 $\pm$ 0.0 & 34.1 $\pm$ 0.0 & 35.4 $\pm$ 0.0 & 54.7 $\pm$ 0.0 & 50.5 $\pm$ 0.0 & 50.3 $\pm$ 0.0 & 44.6 \\
				GFK & 41.4 $\pm$ 0.0 & 26.4 $\pm$ 0.0 & 36.4 $\pm$ 0.0 & 56.2 $\pm$ 0.0 & 43.7 $\pm$ 0.0 & 42.0 $\pm$ 0.0 & 41.0 \\
				CNN & 83.8 $\pm$ 0.3 & 76.1 $\pm$ 0.5 & 80.8 $\pm$ 0.4 & 91.1 $\pm$ 0.2 & 83.1 $\pm$ 0.3 & 89.0 $\pm$ 0.3 & 84.0 \\
				LapCNN & 83.6 $\pm$ 0.6 & 77.8 $\pm$ 0.5 & 80.6 $\pm$ 0.4 & \textbf{92.1} $\pm$ 0.3 & 81.6 $\pm$ 0.4 & 87.8 $\pm$ 0.4 & 83.9 \\
				DDC & 84.3 $\pm$ 0.5 & 76.9 $\pm$ 0.4 & 80.5 $\pm$ 0.2 & 91.3 $\pm$ 0.3 & 85.5 $\pm$ 0.3 & 89.1 $\pm$ 0.3 & 84.6 \\
			  \hline
				DAN$_7$ &\underline{84.7} $\pm$ 0.3 & 78.2 $\pm$ 0.5 & \underline{81.8} $\pm$ 0.3 & 91.6 $\pm$ 0.4 & 87.4 $\pm$ 0.3 & 88.9 $\pm$ 0.5 & 85.4 \\
				DAN$_8$ & 84.4 $\pm$ 0.3 & \underline{80.8} $\pm$ 0.4 & 81.7 $\pm$ 0.2 & 91.7 $\pm$ 0.3 & \underline{90.5} $\pm$ 0.4 & 89.1 $\pm$ 0.4 & \underline{86.4} \\
				DAN$_{\textrm{SK}}$ & 84.1 $\pm$ 0.4 & 79.9 $\pm$ 0.4 & 81.1 $\pm$ 0.5 & 91.4 $\pm$ 0.3 & 86.9 $\pm$ 0.5 & \underline{89.5} $\pm$ 0.3 & 85.5 \\
				DAN & \textbf{86.0} $\pm$ 0.5 & \textbf{81.5} $\pm$ 0.3 & \textbf{82.0} $\pm$ 0.4 & \underline{92.0} $\pm$ 0.3 & \textbf{92.0} $\pm$ 0.4 & \textbf{90.5} $\pm$ 0.2 & \textbf{87.3} \\
        \Xhline{1pt}
    \end{tabular}
\end{table*}

\subsection{Results and Discussion}
The unsupervised adaptation results on the first six \emph{Office-31} transfer tasks are shown in Table~\ref{table:office31}, and the results on the other six \emph{Office-10 + Caltech-10} transfer tasks are shown in Table~\ref{table:office10}. To directly compare with DDC, we report semi-supervised adaptation results of the same tasks used by DDC in Table~\ref{table:semi}. We can observe that DAN significantly outperforms the comparison methods on most transfer tasks, and achieves comparable performance on the easy transfer tasks, \textbf{D} $\rightarrow$ \textbf{W} and \textbf{W} $\rightarrow$ \textbf{D}, where source and target are similar \cite{cite:ECCV10SGF}. This is reasonable as the adaptability may vary across different transfer tasks. The performance boost demonstrates that our architecture of multi-layer adaptation via multi-kernel MMD is able to transfer pre-trained deep models across different domains. 

\begin{table}[tbp]
    \addtolength{\tabcolsep}{-1.8pt}
    \centering
    \caption{Accuracy on \emph{Office-31} dataset with classic unsupervised and semi-supervised adaptation protocols \cite{cite:ECCV10SGF}.}
    \label{table:semi}
    \begin{tabular}{cccccccc}
        \Xhline{1pt}
        Method & A $\rightarrow$ W & D $\rightarrow$ W & W $\rightarrow$ D & Average \\
        \hline
				DDC & 59.4 $\pm$ 0.8 & 92.5 $\pm$ 0.3 & 91.7 $\pm$ 0.8 & 81.2\\
				DAN & \textbf{66.0} $\pm$  0.4 & \textbf{93.5} $\pm$ 0.2 & \textbf{95.3} $\pm$ 0.3 &\textbf{84.9} \\
				\hline
				DDC & 84.1 $\pm$ 0.6 & 95.4 $\pm$ 0.4 & 96.3 $\pm$ 0.3 & 91.9\\
				DAN & \textbf{85.7} $\pm$ 0.3 & \textbf{97.2} $\pm$ 0.2 & \textbf{96.4} $\pm$ 0.2 & \textbf{93.1} \\
        \Xhline{1pt}
    \end{tabular}
\end{table}

From the experimental results, we can make the following observations. (1) Deep learning based methods significantly outperform conventional \emph{shallow} transfer learning methods by a large margin. (2) Among the deep learning methods, the semi-supervised LapCNN provides no improvement over CNN, suggesting that the challenge of domain discrepancy cannot be readily bridged by semi-supervised learning. (3) DDC, a cross-domain variant of CNN with single-layer adaptation via single-kernel MMD, generally outperforms CNN, confirming its effectiveness in learning transferable features using domain-adaptive deep models. Note that while DDC based on Caffe AlexNet was shown to significantly outperform DeCAF \cite{cite:ICML14DeCAF} in which fine-tuning was not carried out, it does not yield a large gain over Caffe AlexNet using fine-tuning. This shows the limitation of single-layer adaptation via single-kernel MMD, which cannot explore the strengths of deep networks and multiple kernels for domain adaptation.

\begin{figure*}[tbp]
  \centering
  \subfigure[DDC Features on Source]{
    \includegraphics[width=0.23\textwidth]{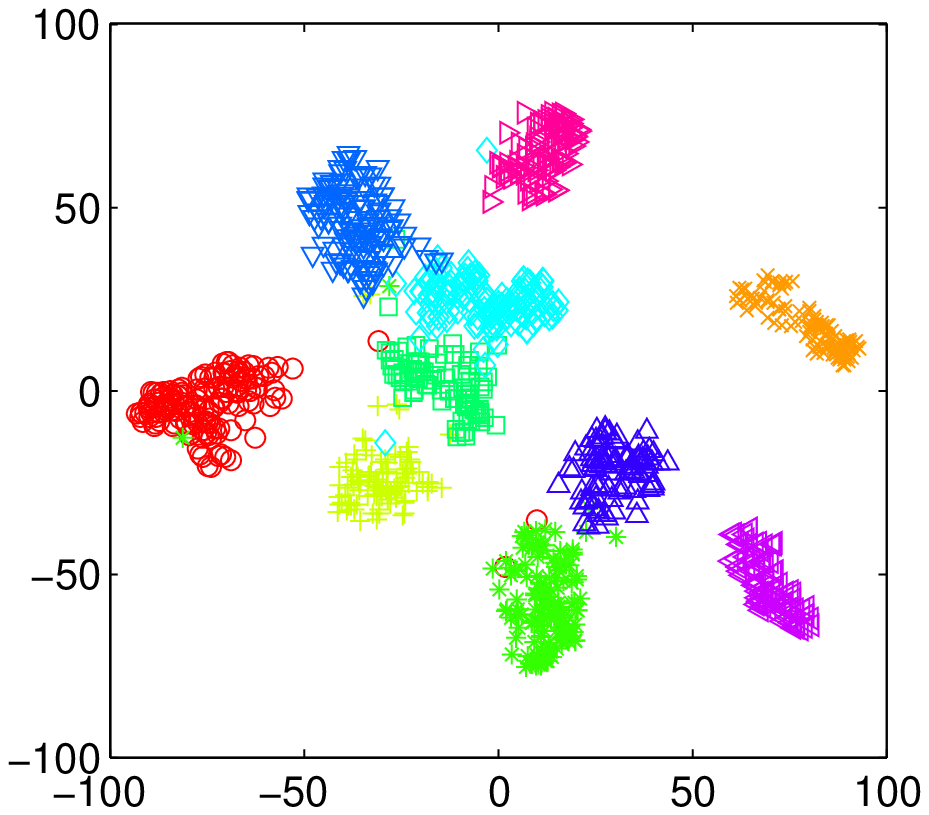}
    \label{fig:CNNs}
  }
  \subfigure[DDC Features on Target]{
    \includegraphics[width=0.23\textwidth]{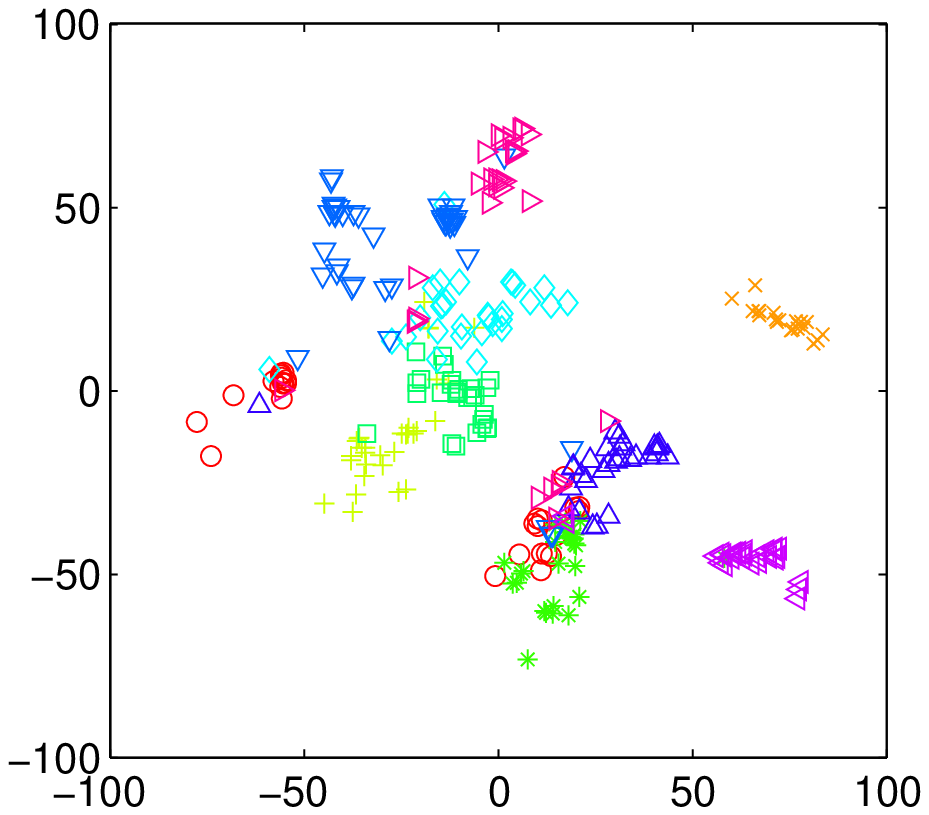}
    \label{fig:CNNt}
  }
  \subfigure[DAN Features on Source]{
    \includegraphics[width=0.23\textwidth]{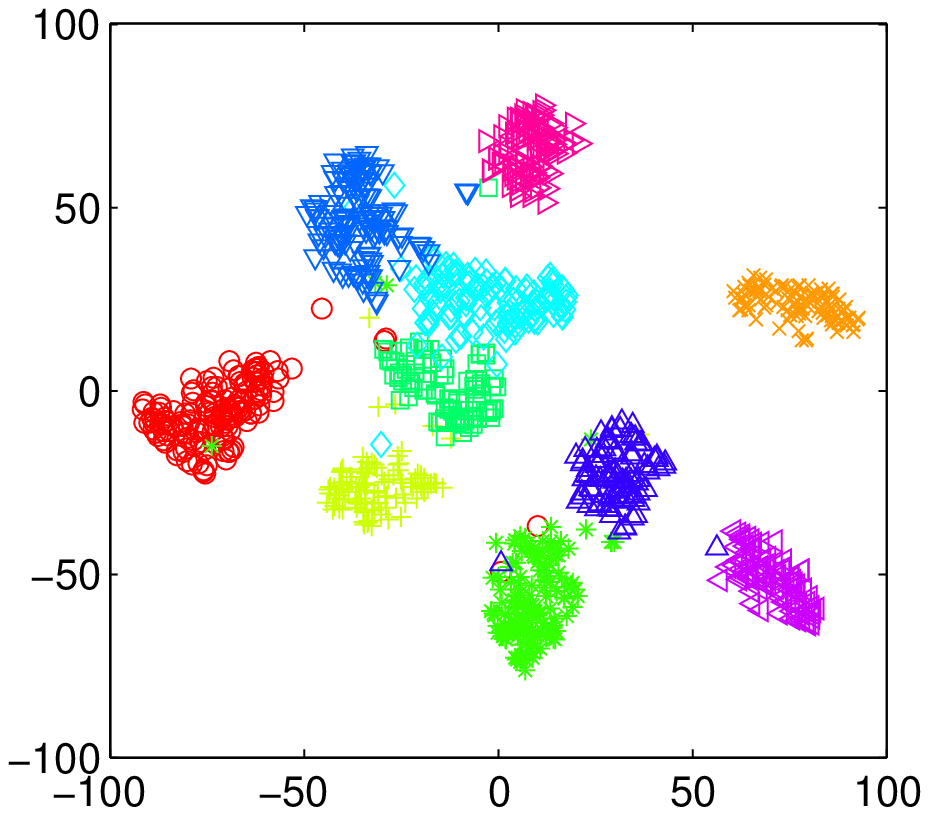}
    \label{fig:DANs}
  }
  \subfigure[DAN Features on Target]{
    \includegraphics[width=0.23\textwidth]{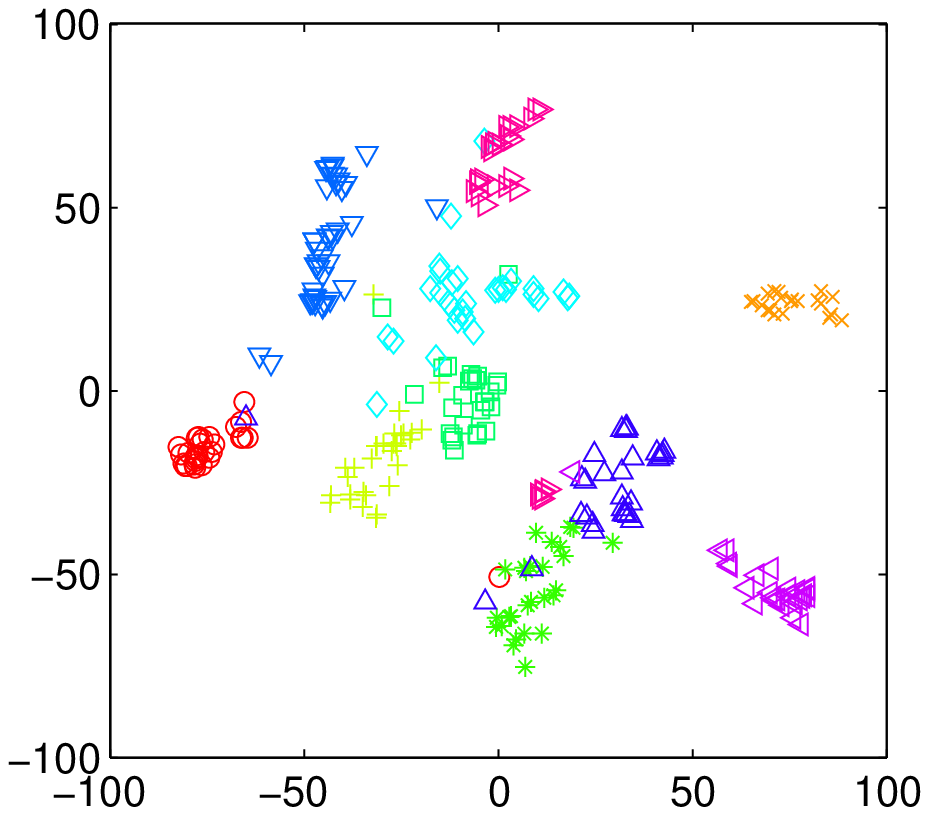}
    \label{fig:DANt}
  }
  \caption{Feature visualization: t-SNE of DDC features on source (a) and target (b); t-SNE of DAN features on source (c) and target (d).}
\end{figure*}

To dive deeper into DAN, we present the results of three variants of DAN: (1) DAN$_7$ and DAN$_8$ achieve better accuracy than DDC, which highlights that multi-kernel MMD can bridge the domain discrepancy more effectively than single-kernel MMD. The reason is that multiple kernels with different bandwidths can match both the low-order moments and high-order moments to minimize the Type II error \cite{cite:NIPS12MKMMD}. (2) DAN$_{\textrm{SK}}$ also attains higher accuracy than DDC, which confirms the capability of deep architecture for distribution adaptation. The rationale is similar to that of deep networks: each layer of deep network is intended to extract features at a different abstraction level, and hence we need to match the distributions at each task-specific layer to consolidate the adaptation quality at all levels. The multi-layer architecture is one of the most critical contributors to the efficacy of deep learning, and we believe it is also important for MMD-based adaptation. The evidence of comparable performance between the multi-layer variant DAN$_{\textrm{SK}}$ and multi-kernel variants DAN$_7$ and DAN$_8$ shows their equal importance for domain adaptation. As expected, DAN obtains the best performance by jointly exploring multi-layer adaptation with multi-kernel MMD. Another benefit of DAN is that it uses a linear-time unbiased estimate of the kernel embedding, which makes it an order more efficient than existing methods TCA and DDC. Though Tzeng et al.~\yrcite{cite:Arxiv14DDC} speed up DDC by computing the MMD within each mini-batch of the SGD, this leads to a biased estimate of MMD and lower adaptation accuracy.

\begin{figure}[tbp]
  \centering
  \subfigure[$\mathcal{A}$-Distance]{
    \includegraphics[width=0.47\columnwidth]{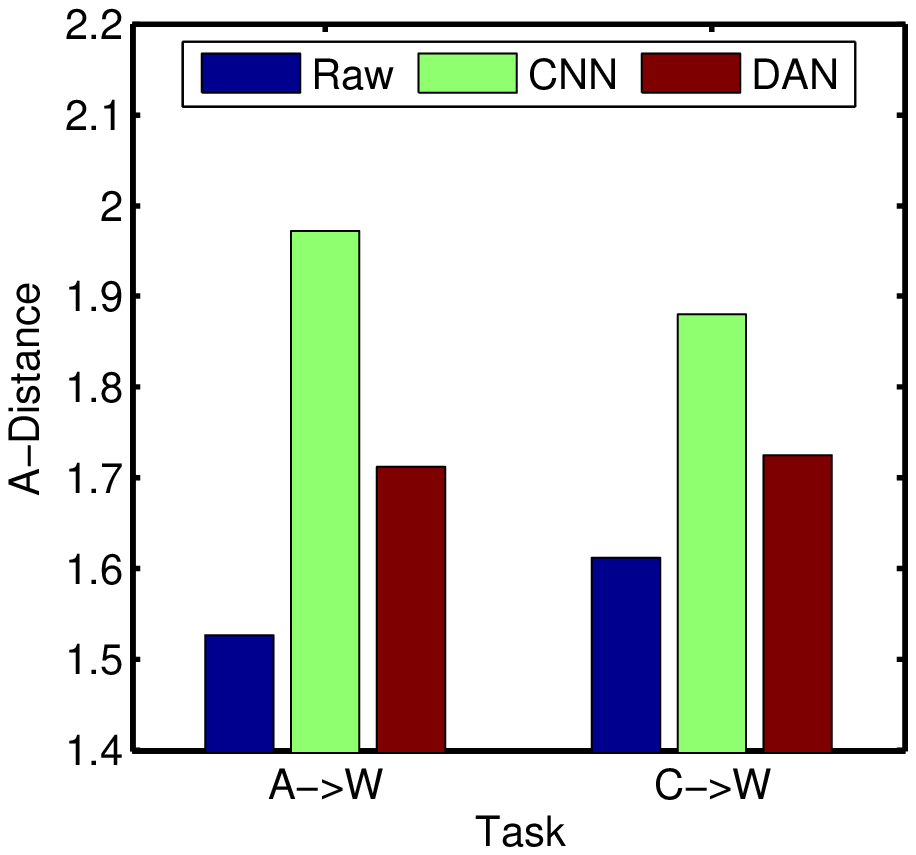}
    \label{fig:Adistance}
  }
  \subfigure[Accuracy vs. $\lambda$]{
    \includegraphics[width=0.47\columnwidth]{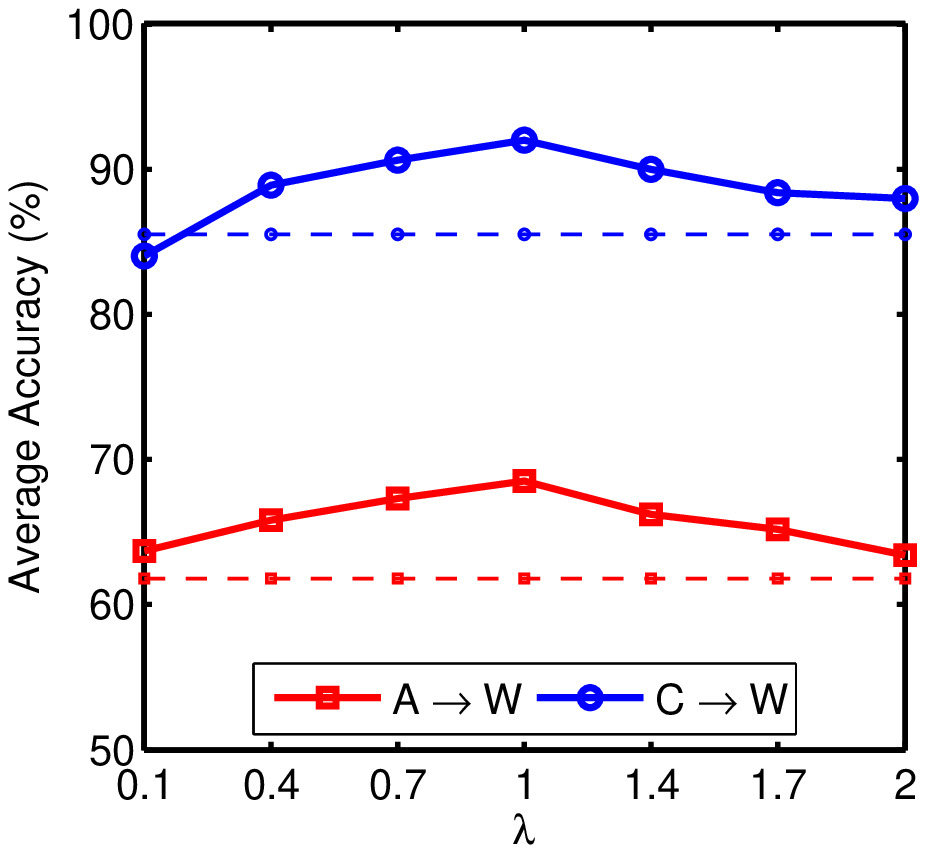}
    \label{fig:Sensitivity}
  }
  \caption{Empirical analysis: (a) $\mathcal{A}$-Distance of CNN \& DAN features; (b) sensitivity of $\lambda$ (dashed lines show best baseline results).}
\end{figure}

\subsection{Empirical Analysis}
\textbf{Feature Visualization} \;
To demonstrate the transferability of the DAN learned features, we follow Donahue et al.~\yrcite{cite:ICML14DeCAF} and Tzeng et al.~\yrcite{cite:Arxiv14DDC} and plot in Figures~\ref{fig:CNNs}--\ref{fig:CNNt} and \ref{fig:DANs}--\ref{fig:DANt} the t-SNE embeddings of the images in task \textbf{C} $\rightarrow$ \textbf{W} with DDC features and DAN features, respectively. We make the following observations: (1) With DDC features, the target points are not discriminated very well, while with DAN features, the points are discriminated much better. (2) With DDC features, the categories between the source and the target are not aligned very well, while with DAN features, the categories are aligned much better between domains. Both these observations can explain the superior performance of DAN over DDC: (1) implies that the target points are more easily discriminated with DAN features, and (2) implies that the target points can be better discriminated with the source classifier. DAN can learn more transferable features for effective domain adaptation.

\textbf{$\mathcal{A}$-Distance} \;
A theoretical result in Ben-David et al.~\yrcite{cite:ML10DAT} suggests $\mathcal{A}$-distance as a measure of domain discrepancy. As computing the exact $\mathcal{A}$-distance is intractable, an approximate distance is defined as ${\hat d_{\cal A}} = 2\left( {1 - 2\epsilon } \right)$, where $\epsilon$ is the generalization error of a two-sample classifier (kernel SVM in our case) trained on the binary problem to distinguish input samples between the source and target domains. Figure~\ref{fig:Adistance} displays ${\hat d}_{\cal A}$ on transfer tasks \textbf{A} $\rightarrow$ \textbf{W} and \textbf{C} $\rightarrow$ \textbf{W} using Raw features, CNN features, and DAN features,  respectively. It reveals a surprising observation that the ${\hat d}_{\cal A}$ on both CNN and DAN features are larger than the ${\hat d}_{\cal A}$ on Raw features. This implies that abstract deep features can be salient both for discriminating different categories and different domains, which is consistent with Glorot et al.~\yrcite{cite:ICML11DADL}. However, domain adaptation may be deteriorated by the enlarged domain discrepancy \cite{cite:ML10DAT}. It is desirable that ${\hat d}_{\cal A}$ on DAN feature is smaller than ${\hat d}_{\cal A}$ on CNN feature, which guarantees more transferable features.

\textbf{Parameter Sensitivity} \;
We investigate the effects of the parameter $\lambda$. Figure~\ref{fig:Sensitivity} gives an illustration of the variation of transfer classification performance as $\lambda \in \{ 0.1, 0.4, 0.7, 1, 1.4, 1.7, 2 \}$ on tasks \textbf{A} $\rightarrow$ \textbf{W} and \textbf{C} $\rightarrow$ \textbf{W}. We can observe that the DAN accuracy first increases and then decreases as $\lambda$ varies and demonstrates a bell-shaped curve. This confirms the motivation of jointly learning deep features and adapting distribution discrepancy, since a good trade-off between them can enhance feature transferability.

\section{Conclusion}\label{section:Conclusion}
In this paper, we have proposed a novel Deep Adaptation Network (DAN) architecture to enhance the transferability of features from task-specific layers of the neural network. We confirm that while general features can generalize well to a novel task, specific features tailored to an original task cannot bridge the domain discrepancy effectively. We show that feature transferability can be enhanced substantially by mean-embedding matching of the multi-layer representations across domains in a reproducing kernel Hilbert space. An optimal multi-kernel selection strategy further improves the embedding matching effectiveness, while an unbiased estimate of the mean embedding naturally leads to a linear-time algorithm that is very desirable for deep learning from large-scale datasets. An extensive empirical evaluation on standard domain adaptation benchmarks demonstrates the efficacy of the proposed model against previous methods.

As deep features transition from general to specific along the network, it is interesting to study the principled way of deciding the boundary of generality and specificity, and the application of distribution adaptation to the convolutional layers of CNN to further enhance the feature transferability.

\section*{Acknowledgments}
This work was supported by the National Science Funds for Distinguished Young Scholars (No. 613250154), National Science and Technology Supporting Program Project (No. 2015BAH14F02), and Tsinghua TNList Fund for Big Data Science and Technology.

\begin{small}
	\bibliography{DAN}
	\bibliographystyle{icml2015}
\end{small}

\end{document}